\documentclass[letterpaper, 10 pt, conference]{ieeeconf}  % Comment this line out if you need a4paper

\IEEEoverridecommandlockouts                              % This command is only needed if 
\usepackage{xcolor}
\usepackage{listings}

\usepackage{algorithm,algpseudocode}
\usepackage{amsmath,amssymb,amsfonts}
\usepackage{graphicx}
\usepackage{caption}
\usepackage{subcaption}
\usepackage{pdfpages}
\usepackage{cite}
\usepackage{multirow}
\usepackage{diagbox}
\colorlet{mygray}{black!30}
\colorlet{mygreen}{green!60!blue}
\colorlet{mymauve}{red!60!blue}

\title{\LARGE \bf
coVoxSLAM: GPU accelerated globally consistent dense SLAM 
}

\author{Emiliano H\"oss and Pablo De Crist\'oforis% <-this % stops a space
\thanks{Both authors are with the Department of Computer Science, Faculty of Exact and Natural Sciences, University of Buenos Aires, Argentina
        {\tt\small ehoss@dc.uba.ar};
        {\tt\small pdecris@dc.uba.ar} }%
        }%

\begin{document}

\maketitle
\thispagestyle{empty}
\pagestyle{empty}

%%%%%%%%%%%%%%%%%%%%%%%%%%%%%%%%%%%%%%%%%%%%%%%%%%%%%%%%%%%%%%%%%%%%%%%%%%%%%%%%
\begin{abstract}
A dense SLAM system is essential for mobile robots, as it provides localization and allows navigation, path planning, obstacle avoidance, and decision-making in unstructured environments. Due to increasing computational demands the use of GPUs in dense SLAM is expanding. 
In this work, we present coVoxSLAM, a novel GPU-accelerated volumetric SLAM system that takes full advantage of the parallel processing power of the GPU to build globally consistent maps even in large-scale environments. It was deployed on different platforms (discrete and embedded GPU) and compared with the state of the art. The results obtained using public datasets show that coVoxSLAM delivers a significant performance improvement considering execution times while maintaining accurate localization. The presented system is available as open-source on GitHub\footnote{https://github.com/lrse-uba/coVoxSLAM}.

\end{abstract}

%%%%%%%%%%%%%%%%%%%%%%%%%%%%%%%%%%%%%%%%%%%%%%%%%%%%%%%%%%%%%%%%%%%%%%%%%%%%%%%%
\section{INTRODUCTION}

In order to achieve the goal of autonomously navigating and interacting in unknown environments, mobile robots need to build a map of the observed world. Since robots estimate their motion incrementally as they move through the environment, this leads to unbounded errors in pose estimation and inconsistent mapping. Thus, maintaining a globally consistent map represents a key functionality in SLAM (Simultaneous Localization and Mapping) for which loop closing is required, as it allows the system to recognize previously visited locations and reduce accumulated errors of both the robot's trajectory and the generated map.

Several well-known SLAM systems, which convert raw sensor data into features for map building, have demonstrated the ability to create globally consistent maps in real time~\cite{mur2017orb, labbe2019rtab, cramariuc2022maplab}. Even so, these feature-based maps are restricted use for tasks beyond localization due to the difficulties of extracting the shape and connectivity of surfaces from a sparse map. 

On the other hand, dense maps are suitable not only for pose estimation but also for path planning, obstacle avoidance, and autonomous navigation. 
Priors dense SLAM systems were developed based on photometric bundle adjustment~\cite{newcombe2011dtam, kerl2013dense, whelan2015elasticfusion}
and most currently using deep learning~\cite{czarnowski2020deepfactors, koestler2022tandem, yang2022vox}. However, these systems are still compute-intensive and struggle to run in real-time in large-scale scenarios.
\begin{figure}[t]
    \centering
	\includegraphics[width=\columnwidth]{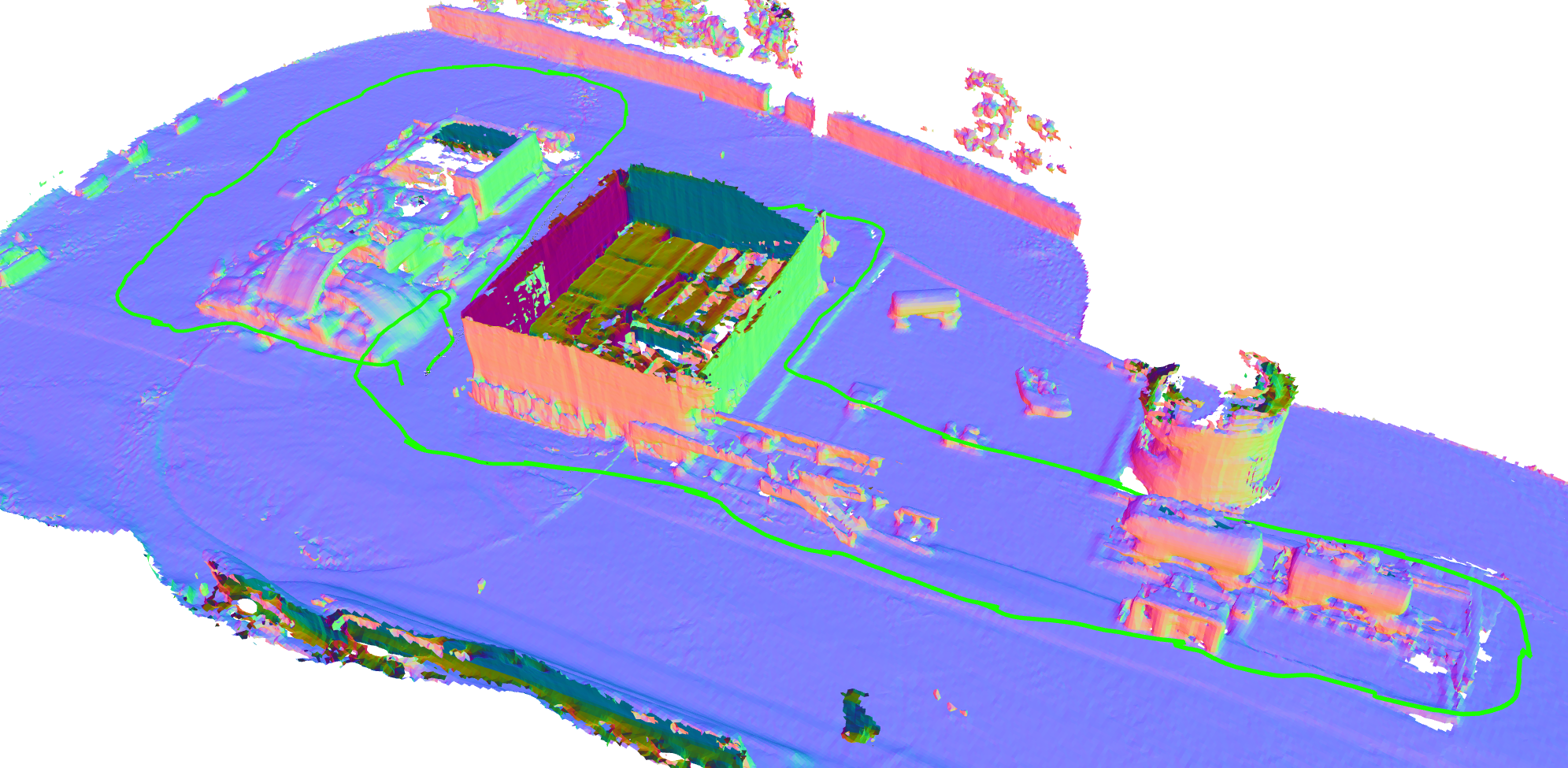}
	\caption{A reconstruction resulting from a 400 m-long MAV flight through the search and rescue training site discussed in Sec. \ref{results}. In the foreground a large pile of rubble from a collapsed structure is visible. The trajectory
(green) also contains indoor-outdoor transitions through a building.}
	\label{fig:coverImage}
\end{figure}

Implicit surfaces modelled as Signed Distance Fields (SDFs), introduced in~\cite{curless1996volumetric}, have been first proven to be an effective representation for dense mapping by KinetFusion~\cite{newcombe2011kinectfusion}. It uses the Truncated Signed Distance Field (TSDF) as a volumetric representation and Fast Iterative Closest Point (ICP) for pose estimation, to provide a real-time fused dense model of the scene. However, it has limitations in an unbounded extended area and tracking failures in environments with poor 3D geometry.

Producing a globally consistent map using SDFs is expensive since global optimization of the map quickly becomes intractable as the amount of data increases.  An alternative to address this issue is representing the reconstructed environment as a collection of submaps. The advantage of this approach is that the sensor pose, at the time of new sensing integration, only needs to be registered with respect to the current submap~\cite{fioraio2015large}. Voxgraph ~\cite{reijgwart2019voxgraph} extends this idea by proposing to use correspondence-free alignment based on the Euclidean Signed Distance Field (ESDF) that represents a voxel grid where every point contains its Euclidean distance to the nearest obstacle. Voxgraph uses Voxblox~\cite{oleynikova2017voxblox} to incrementally build the ESDF directly out of TSDFs, and leverage the distance information already contained within the truncation radius. Another system that builds a global ESDF map incrementally is FIESTA~\cite{han2019fiesta} that proposes an elaborately designed data structures and a novel ESDF updating algorithm.

Building on these advances, we present a novel GPU accelerated globally consistent dense SLAM system, coined coVoxSLAM that outperforms the state of the art considering execution times without compromising the accuracy. The new system can reach real-time even in embedded devices such as those that can be mounted on board small mobile robots.

\section{RELATED WORK}
\begin{figure*}[t]
    \centering
	\includegraphics[width=2\columnwidth]{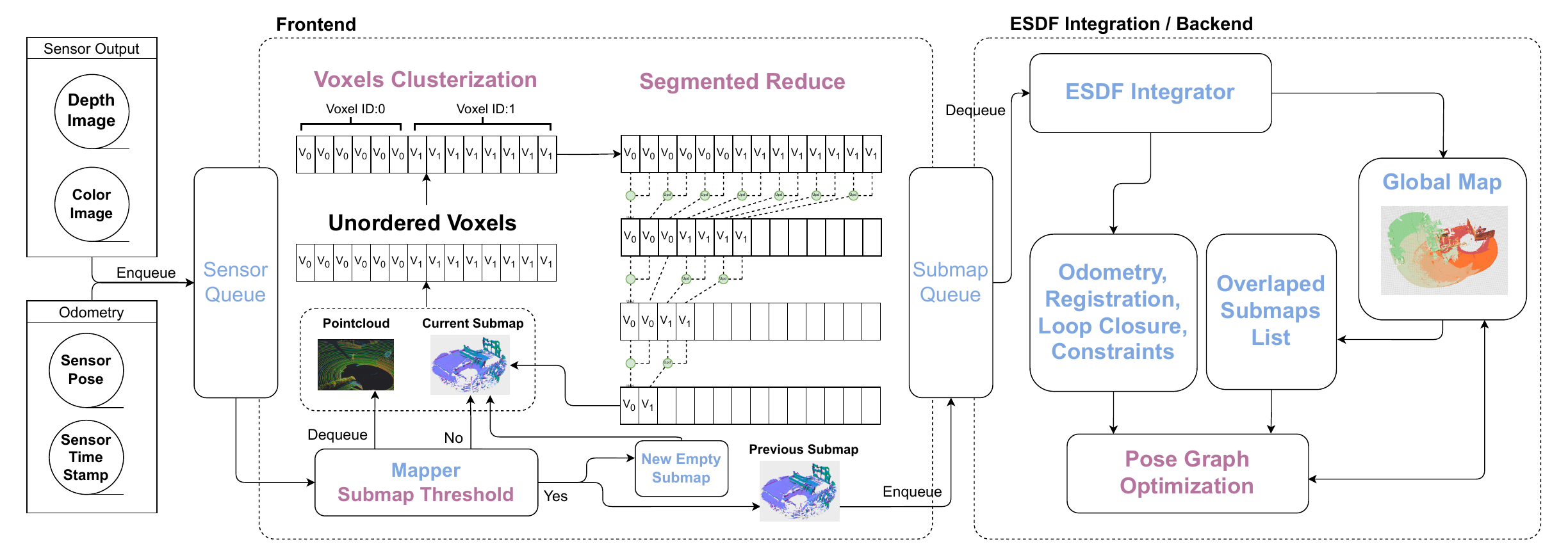}
	\caption{Data flow pipeline of the system, using two queues for the front end and the back end to improve reliability. The mapper reads from the queue when a new pointcloud is available. When the submap threshold is reached, it enqueues the current submap to be finalized by the ESDF Integrator, ends the backend and returns a new empty submap to the TSDF Integrator. Otherwise, it returns the current submap. The size of both queues can be modified as a parameter of the system.}
	\label{fig:pipeline}
\end{figure*}

During the last decade, efforts have been made to implement SLAM algorithms on multiple processing units in heterogeneous CPU-GPU systems. That represents a more realistic approach to their lifelike usage, highlighting the benefits and disadvantages of embedded computing solutions for mobile robots. Nevertheless, porting algorithms originally designed for CPUs to GPUs is not straightforward because of differences in hardware architectures and resources. Such as the case of \cite{aldegheri2019data, giubilato2019evaluation, song2018mis}, where GPU-accelerated versions of features-based SLAM systems are evaluated. The GPU has also been used for some components of dense RGB-D SLAM~\cite{newcombe2011kinectfusion, schops2019bad}. In LiDAR SLAM's context, GPU use was mostly limited to accelerating scan matching in the frontend~\cite{behley2018efficient, koide2021globally}.  However, in all of these previous works, the backend, and specifically, the pose graph optimization, remains performed on a CPU. More recently, nvBlox\cite{millane2024nvblox} presents a library for volumetric mapping with GPU acceleration to achieve high performance and real-time 3D reconstruction. It inherits from Voxblox and focuses on optimizing it for GPU processing to handle TSDF and ESDF map representations, but it does not solve SLAM because it does not consider pose graph optimization neither loop-closing. This paper introduces coVoxSLAM: an entirely (frontend and backend) GPU accelerated SLAM for on-board globally consistent volumetric mapping in real-time that outperforms nvBlox in the TSDF integration achieving a speed-up of $1.5 \times$ to $2\times$ in the same datasets.

\section{SYSTEM ARCHITECTURE DESCRIPTION}
The architecture of coVoxSLAM is shown in Fig.~\ref{fig:pipeline}. The system consists of a frontend and backend. The frontend is responsible for integrating the incoming sensor data into a TSDF volume to include or update the voxels that build the TSDF maps and propagate updated voxels from the TSDF to the ESDF submaps. The created voxels are grouped in fixed-size blocks, which in turn, are indexed using an appropriately designed hash table. The backend is responsible for estimating the most likely submap collection alignment by minimizing the total error of the three pose graph constraints: odometry, loop closure and submap registration. We discuss these components in detail below.

\subsection{Data Layout}

coVoxSLAM introduces a versatile container that allows users to choose between two different data layouts: the Structure of Arrays (SoA) and the opposite Array of Structs (AoS) formats. Unlike the traditional AoS format, where each element is a structure containing multiple fields, the SoA format stores each field in a separate, contiguous array. By storing homogeneous data together, SoA minimizes cache misses and maximizes the use of cache lines, leading to better spatial locality. 
SoA eliminates the need for padding within individual structures, minimizing memory consumption and maximizing memory throughput. 
Moreover, there is a growing disparity between the speed at which CPUs can process data and the speed at which memory can supply data to the CPU. 
Consequently, this layout can significantly boost performance in memory-bound applications, especially those involving large input data and parallel processing.

In contrast to integrating current implementations of SoA containers~\cite{jubertie2018data, homann2018soax, hwu2011gpu, gruber2023llama}, our tailored implementation provides a zero-cost abstraction layer, meaning it incurs no runtime overhead and seamlessly integrates with existing codebases. It leverages automatic vectorization to fully exploit SIMD (Single Instruction, Multiple Data) capabilities. 
According to our experiments, SoA's performance improves in look-up query execution times, which results proportional to the size of the queried type relative to the total size of the main structure.

\subsection{Data Structures}

Space-partitioning data structures organize and manage spatial data enabling rapid querying, insertion, and deletion operations. %Trees and hash tables are particularly prominent in fields such as Simultaneous Localization and Mapping (SLAM) and Computer Graphics. % among the various types of space-partitioning data structures. 
Tree-based structures, such as Octomaps~\cite{hornung2013octomap}, BVH-trees~\cite{lauterbach2009fast}, and KD-trees~\cite{bentley1990k}, are widely used for their ability to partition space hierarchically, making them ideal for tasks like collision detection, nearest neighbour search, and mapping. However, most of the GPU-accelerated versions of both KD-trees \cite{rusu20113d, zhou2008kdtree} and BVH ~\cite{wald2007ray, lauterbach2009fast, gunther2007realtime} do not allow for dynamic-sizing since allocating memory does not scale on GPU. 

%These data structures are critical in optimizing performance and ensuring scalability in complex systems, making them indispensable in modern robotics, SLAM, and computer graphics applications.

%alternatives
%Trees, like KD-trees, BVHs, and Octomaps, are data structures built by recursively partitioning space.  
%KD-tree~\cite{bentley1975kdtree} is particularly useful for applications involving multidimensional search keys, such as range searches and nearest neighbour searches. Each node in a KD-tree represents a k-dimensional point. 
%Non-leaf nodes act as hyperplanes that divide the space into two half-spaces. 
%Bounding volume hierarchy (BVH) is another hierarchical representation that organizes 2D or 3D primitives. Each node in the BVH represents a bounding volume that encloses a subset of the objects. It’s instrumental in computer graphics, especially for collision detection and ray tracing tasks.

%This behaviour originates from the space-splitting technique based on data distribution. 
%Another data structure, the Octree~\cite{meagher1982octree}, bases its splitting on recursively partitioning the 3D space into eight equivalent subvolumes conditionally when an object is occupying. It has been widely used in adaptive 3D mapping~\cite{hornung2013octomap, museth2013openvdb} for robot navigation ~\cite{dong2022ash}. 

%definitions
On the other hand, hash tables are data structures used to store \textit{sparse keys}, providing efficient indexing and retrieval of spatial data through hashing techniques.
%$(e.g.\ unbounded indices, strings, coordinates) from the set $\mathcal{K}$ to values from the set $\mathcal{V}$ because of their compact data layout and expected constant time complexity for insertion and retrieval. 
%It uses a hash function $h: \mathcal{K} \to \mathcal{I}_n$ where $\mathcal{I}_n = \{0, 1, \dotsc, n-1\}$ to minimize the number of collisions and improve the distribution of values according to arbitrary criteria.
%state-of-art
Despite the hash function, a collision resolution scheme is still needed. Most GPU hash tables use open addressing~\cite{alcantara2009real, junger2020warpcore, garcia2011coherent} because of the limitations in GPU to work with dynamic-sized structures and the possibility to implement highly concurrent operations.
While hash map implementations are widely available for CPU, their GPU counterparts have only emerged in the recent decade~\cite{dong2022ash}.

%Spatial hashing is another variation of \textit{spatial management} with $O(1)$ access time depending on hash maps. Bundled with small dense 3D arrays, it has been widely used in real-time volumetric scene reconstruction within \emph{unbounded} region of interest.

%related
Some implementations include stdgpu~\cite{stotko2019stdgpu} but are not optimized for large input data. Recently, WarpCore~\cite{junger2020warpcore} proposes to support non-integer values and dynamic insertion, but the key domain is still limited to, at most, 64-bit integers. ASH~\cite{dong2022ash} provides a device-agnostic dynamic, generic and collision-free hash table, an improved SoA data layout and several downstream applications that achieve higher performance than state-of-the-art implementations, including TSDF reconstructions.

In coVoxSLAM, a similar approach to the ASH hash table is adopted: buffers are utilized in a SoA layout to enhance data processing efficiency. In addition for \textit{insert} and \textit{find} operations, ASH introduce a new functionality that \textit{activates} the input keys by inserting them into the hash map and obtaining the associated buffer indices. We make use of this ASH feature to accelerate the allocation of blocks in our multi-step process for the TSDF integration in the frontend. 

\subsection{TSDF Representation}
\label{mapping}

Map representation will significantly impact the system's performance and accuracy. Occupancy maps use a grid of cells (voxels in 3D) where each cell holds a probability value indicating whether it is occupied, free, or unknown~\cite{moravec1985high}. TSDF maps store the distance to the nearest surface within a certain truncation distance. The main advantage of TSDF over occupancy maps is smoothing noise after successive observations~\cite{curless1996volumetric}.

Using one hash table to discretize the entire environment might not scale well when either the size of the map or the resolution increases, and it could also maximize cache misses since neighbourhood voxels are scattered in the hash table. The current state-of-art volumetric discretization for TSDF utilizes two levels of hierarchy~\cite{oleynikova2017voxblox, millane2024nvblox}, representing the explored environment, where small dense grids are the minimal unit in the hash table. Recent works ~\cite{millane2018c, reijgwart2019voxgraph} extend this idea by representing the global map as a combination of overlapping submaps alongside its position and orientation to achieve global consistency. Each submap uses a sparse collection of blocks, aiming to represent only the parts of the environment where it has gathered sensor data.

In this work, the submaps and blocks are indexed by their 3D position in the map. The returned index is used to access data in memory. There is no need for a double indirection looking for the pointer in a secondary hash table. The data from voxels, such as weight or distance, or any other data declared in a user-defined struct, is densely packed in memory using the SoA container which maximizes the number of coalesced access in GPU memory and also facilitates the synchronization of memory writes in non-coherent memory.

As described in~\cite{oleynikova2017voxblox}, the choice of weighting function can strongly impact the accuracy of the resulting reconstruction, for large voxels or surfaces close to the sensor, where thousands of points may be merged into the same voxel, and also for distant voxels where sensor quality may be degraded. Different weights should be assigned depending on the performance curve of the sensor. However, not every map representation and its construction methods allow the use of different weighting techniques. For example, nvBlox uses an interpolation technique during weighting to approximate the confidence level of each voxel's data that results in a loss of accuracy.

Regarding how sensor data is merged into the TSDF map, there are two main methods: raycasting~\cite{oleynikova2017voxblox} and projection mapping ~\cite{newcombe2011kinectfusion, klingensmith2015chisel}. Raycasting casts rays from the camera optical centre to the centre of each observed point and updates all voxels from the centre to truncation distance behind the point. The main drawback of this technique is that it is challenging to implement on GPUs. Projection mapping instead projects voxels in the visual field of view into the depth image and computes their distance from the difference between the voxel centre and the depth value in the image. 
This might lead to strong aliasing effects for larger voxels, incomplete reconstructions and loss of essential details.
Moreover, projecting each voxel into the camera image and associating it with the nearest pixel can be computationally intensive, especially for high-resolution grids. Projection methods can be sensitive to sensor noise, which can introduce errors into the TSDF~\cite{werner2014truncated, klingensmith2015chisel, DTSDF_IROS_2019, klingensmith2015chisel}
To overcome these limitations, coVoxSLAM does not use projection mapping, but instead takes advantage of our improved GPU raycasting implementation, which considers every point from the sensor when evaluating a voxel, facilitating the integration of different weighting techniques. This is one of the main differences of our system compared to nvBlox.

\subsection{Point cloud Integration}
\label{tsdf}

New submaps are generated at regular intervals based on elapsed time or distance travelled, preventing odometry errors from accumulating excessively. 
In coVoxSLAM, the area covered by sensor trajectory is approximated by considering the number of blocks used by the current submap. This approximation is computationally cheap and maximizes consistency in the number of blocks for each submap.

coVoxSLAM employs a multi-step approach. 
During the preliminary counting phase, the algorithm counts the new voxels and blocks.
This phase is highly parallelizable because each thread can independently process different segments of the sensor data. 
Two lists of new and unique voxels are also created during this step. 
Counting is typically discouraged in GPU contexts because it can lead to contention issues. 
Instead of relying on a single global counter, our method uses reduction techniques where each thread or block maintains its local counter, and the results are combined at the end.

During the memory allocation phase, a pre-allocated buffer handles new voxels and blocks. Threads request portions of this buffer and increment their indices using atomic operations.

During the processing phase, the system updates the TSDF's values for the unique voxels. Because the calculations for weight and distance are commutative and distributive, the segmented reduce method is used to distribute the workload among GPU threads and merge intermediate results incrementally. Additionally, the algorithm sorts the voxels by their memory addresses to prevent race conditions when writing the final results to the grid. This can be seen with more detail in the Frontend of Fig.~\ref{fig:pipeline}.

Each GPU thread handles a similar amount of work, focusing only on the voxels intersected by a ray from the sensor. The system maximizes the GPU’s efficiency by equalising the workload, keeping more threads active and effectively utilized. This balance ensures that no thread is overworked while others are idle. This results in better overall performance and resource use, as the GPU can process more data in parallel without delays caused by uneven work distribution.

As the resolution of the map increases, projecting voxels from a volume grid into the sensor becomes less efficient. This can be notice in Fig. 6 from nvBlox~\cite{millane2024nvblox} where the speed-up between nvBlox and Voxblox decreases as the resolution increases. The reason is that blocks (called \textit{VoxelBlocks} in nvBlox) are not small enough to fit only the area intersected by the sensor’s rays, and instead, they form cubic-like volumes. Therefore, as the resolution increases, the number of voxels grows almost cubically.
Moreover, in the most extreme case, when blocks are small enough that each block intersects only one ray, there are no benefits to casting through the block's grid.
Additionally, the operations required to re-project the voxels from the grid, back to the sensor image are computationally more expensive than those involved in ray-casting. This further emphasizes the inefficiency of projection mapping compared to raycasting, which directly processes the intersected voxels, leading to better performance and scalability.

\subsection{ESDF computation}

ESDF maps help path planning algorithms run faster by providing quick access to distance and gradient information against obstacles. However, the challenge lies in keeping these maps updated in real time when the environment changes. We base our ESDF computation on the Parallel Banding Algorithm (PBA)~\cite{cao2010parallel}.

Using threads to handle each voxel or a block can theoretically increase parallelism because they can work simultaneously on different parts of the map, potentially speeding up the computation. 
However, this approach can lead to a completely random pattern of write operations. Each thread might try to write to different parts of the memory simultaneously, causing conflicts and race conditions. 
Memory access patterns must be optimised to achieve better performance. This might involve organizing the data to minimize conflicts. 

In coVoxSLAM, after raycasting, the map is updated with all newly observed voxels (i.e., previously unknown but now detected as occupied or free). Every time a voxel goes from occupied to free, the ESDF integrator needs to propagate around its neighbour voxels. Conversely, when a voxel goes from free to occupied, the calculation of updates is computed around its neighbour voxels to change old distances with potential new values when the distance to the newly occupied voxel is smaller.

When a voxel updates its distance, it must also propagate this new information to its neighbours. This is the third case when more voxels may need to be queued for processing. We subdivide the ESDF map into regions using one block per region. Blocks are dense matrices of $n^3$ voxels, typically $n=8$ or $n=16$. Each warp in the GPU updates one block at a time. These regions are chosen by their index, so every block in each region is contiguous in memory. Each region uses its own queue, and every time a newly occupied or free voxel needs to queue its neighbours voxels, it does it in the queue of the region where they belong. 

Similar to nvBlox, we process every axis direction within each queued block in parallel. After that, the boundaries of every neighbour block are updated if they register a value from a more distant surface. This process is repeated until no more voxels meet the mentioned condition.

\subsection{Backend}
\label{backend}

The backend aligns the submap collection by minimizing the total error of three pose graph constraints: odometry, loop closure and submap registration. The pose graph optimization (PGO)~\cite{besl1992method} consists of the estimation of a set of poses (positions and orientations) from relative pose measurements. %The problem can be formulated as a nonconvex minimization and visualized as a graph. 
To do that, we solve the non-linear least square minimization following 
\cite{reijgwart2019voxgraph}
\begin{equation} 
\begin{aligned}
\underset{\mathcal{X}}{\mathrm{arg~min}} & \sum_{(i,j) \in \mathcal{R}} || e_{\mathrm{reg}}^{i,j}(T_{WS^i},T_{WS^j}) ||_{\sigma_{\mathcal{R}}}^2 + \\
& \sum_{(i,j) \in \mathcal{O}} || \boldsymbol{e}_{\mathrm{odom}}^{i,j}(T_{WS^i},T_{WS^j}) ||_{\Sigma_{\mathcal{O}}}^2 + \\
& \sum_{(i,j) \in \mathcal{L}} || \boldsymbol{e}_{\mathrm{loop}}^{i,j}(T_{WS^i},T_{WS^j}) ||_{\Sigma_{\mathcal{L}}}^2 
\end{aligned}
\label{eq:backend}
\end{equation}
where
\begin{equation*}
\mathcal{X} = \{T_{WS^1}, T_{WS^2},\dots,T_{WS^N} \} 
\end{equation*}
are the submap poses, $T_{WS^i} \in \mathbb{R} \times SO(2)$, and $\mathcal{O}$, $\mathcal{R}$ y $\mathcal{L}$ are sets containing submap index pairs joined by odometry, registration and loop-closure constraints respectively. For more details about the formulation of the problem, you can read~\cite{reijgwart2019voxgraph}.

We use ICP to measure the distance between every overlapped submap in the collection, which involves minimizing the Euclidean distance between the points of each submap. Every time a new submap is created, it could potentially influence the position and orientation of any other submap and even trigger the fuse between two pre-existing submaps. Calculating overlapping submaps is approximated using Axis Aligned Bounding Boxes (AABBs).

During the first step, the ICP algorithm's relevant points are arbitrarily taken from each submap based on its weights. The second step of the backend is related to the least square minimization problem. Instead of using a computationally expensive nonlinear minimization library like Ceres~\cite{Agarwal_Ceres_Solver_2022}, we developed a lightweight and minimal GPU accelerated implementation to solve the equation Eq.~\eqref{eq:backend}.

We use the Conjugate Gradient (CG) method to solve the reduced system iteratively. CG is well-suited for large, sparse systems and benefits from good preconditioning. The Levenberg-Marquardt (LM) algorithm iteratively updates the parameters to reduce the sum of squared residuals using the Trust Region method for searching the step that minimizes the objective function. Both the Jacobians and the Residuals are pre-calculated and pre-allocated in the GPU. In both steps, the calculation of the registration points from the global map until the least square minimization problem using an iterative solver is computed entirely on the GPU.

\section{EXPERIMENTS AND RESULTS}
\label{results}
In this section we validate our claims using two distinctive datasets: four outdoor large-scale datasets used in the same public domain\footnote{Available at: http://robotics.ethz.ch/$\sim$asl-datasets/2020\_voxgraph\_arche} presented alongside Voxgraph and two indoor synthetic room-scale datasets that are used in nvBlox, Replica~\cite{straub2019replica} and Redwood~\cite{park2017colored}. The first four datasets correspond to flights carried out by a Micro Aerial Vehicle (MAV) hexacopter mounted with an Ouster OS1 LiDAR, each with approx—400 m long trajectory around a disaster area designed for training rescuers. The ground truth was generated from an RTK-GNSS system.

The processing units used to perform the experiments were a standard desktop PC (CPU with an AMD Ryzen 9 5950x processor and an Nvidia GeForce RTX 2060 GPU) and the Nvidia Jetson Xavier AGX board. 

\begin{figure}[t]
    \centering
	\includegraphics[width=.75\linewidth]{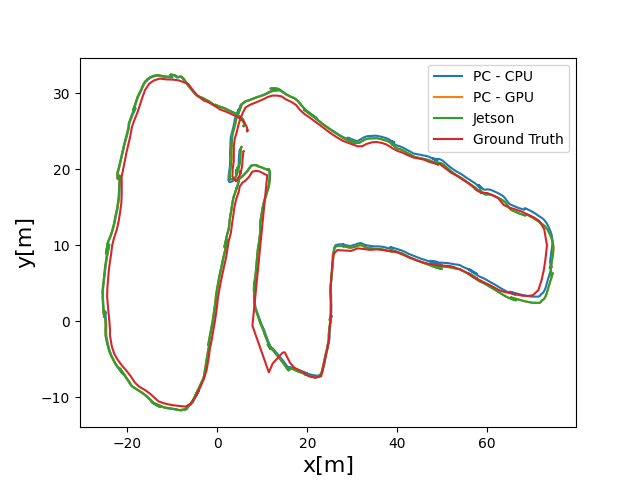}
	\caption{The estimated trajectories of Flight 1 dataset by Voxgraph (blue), coVoxSLAM on PC (orange), coVoxSLAM on Jetson Xavier AGX (green) and RTK-GNSS measurements (red) used as ground truth to evaluate the system.}
	\label{fig:hashtable}
\end{figure}

\subsubsection{TSDF Timings}

\begin{figure*}[h]
\begin{subfigure}{.5\textwidth}
    \centering
	\includegraphics[width=.75\columnwidth]{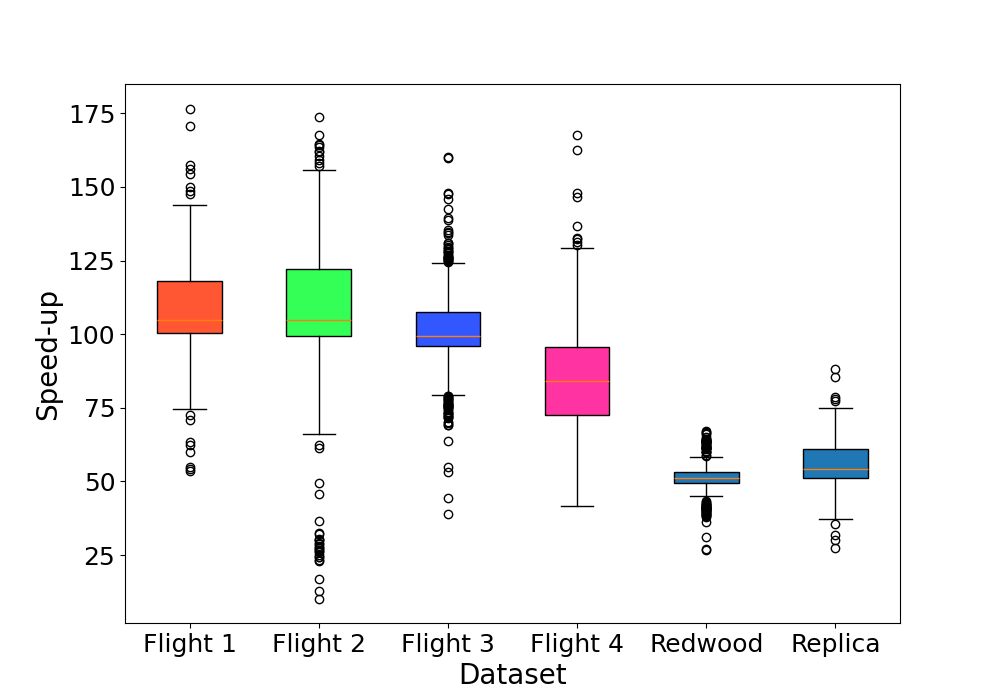}
	\caption{Speep-up of execution times of TSDF integration in six different datasets.}
	\label{fig:tsdf_timing}
\end{subfigure}
\begin{subfigure}{.5\textwidth}
    \centering
	\includegraphics[width=.75\columnwidth]{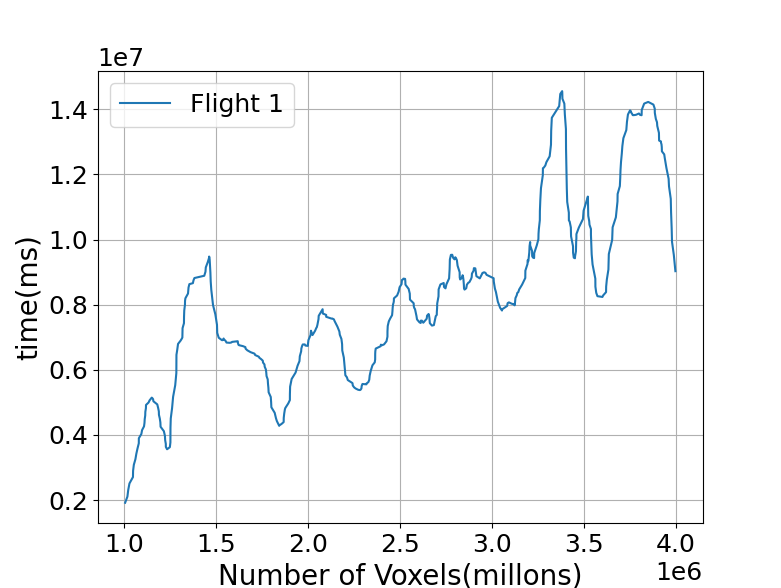}
	\caption{Execution time for Flight 1 dataset, increasing the number of voxels from one to four million.}
	\label{fig:tsdf_by_voxels_count}
\end{subfigure}
\caption{Execution times for the TSDF Integrator, including TSDF + Color.}
\end{figure*}

We first compare coVoxSLAM against Voxblox for the TSDF integration module of both TSDF and Color. Fig.~\ref{fig:tsdf_timing} shows the speed-up for several point clouds. Flights 1 to 4 are four large-scale maps; they show a speed-up between $30\times$ to $140\times$ with an average of $100\times$. The last two datasets, Replica and Redwood, are small indoor datasets; they present a speed-up of $50\times$ on average. This represents an increment of $2\times$ and $1.5\times$ against nvBlox, which reports $38\times$ and $25\times$ against Voxblox for these last two datasets.
This evidences that using raycasting to integrate TSDF outperforms projection mapping used in nvBlox. This is considering TSDF + Color integration. Finally, Fig. ~\ref{fig:tsdf_by_voxels_count} shows how the method scales when incrementing the number of voxels. We aim to show that raycasting methods scale almost linearly when the number of voxels increases.

\subsubsection{ESDF Timings}

\begin{figure}[t]
\centering
	\includegraphics[width=.75\columnwidth]{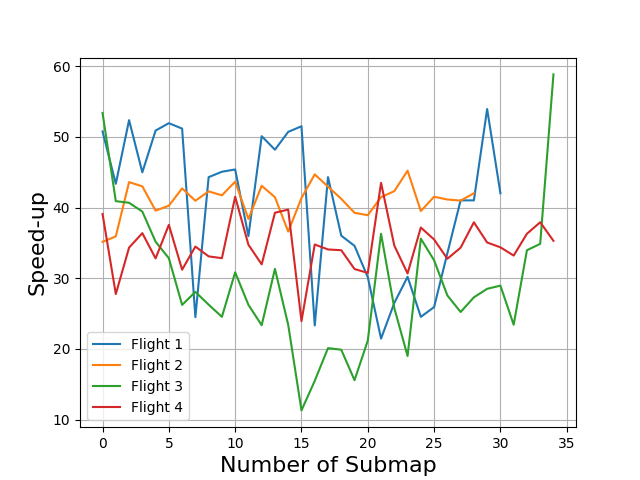}
	\caption{Speep-up of execution times of ESDF integration in four different datasets.}
	\label{fig:esdf_timing}
\end{figure}

We evaluate coVoxSLAM against Voxgraph for the ESDF integration module. Fig.~\ref{fig:esdf_timing} shows the speed-up for several point clouds integrated in four datasets. It shows a speed-up between $10\times$ and $50\times$ of increment.

\subsubsection{Backend Timings}

\begin{figure}[t]
\centering
	\includegraphics[width=.75\columnwidth]{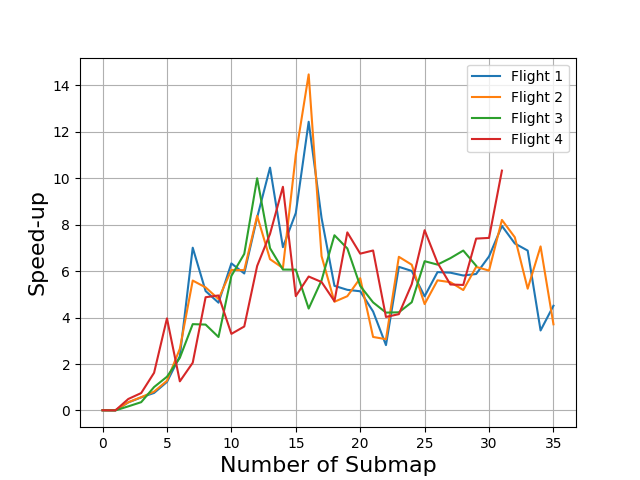}
	\caption{Speep-up of execution times of the Backend in four different datasets.}
	\label{fig:backend_timing}
\end{figure}

We compute coVoxSLAM against Voxgraph, which uses Ceres on the CPU to compute the pose graph optimization process. Fig.~\ref{fig:backend_timing} shows the speed-up between coVoxSLAM and Voxgraph implementations for several executions in four datasets. It shows a speed-up between $2\times$ and $14\times$ of increment. 

\subsubsection{Hash table stress testing}

We use a synthetic dataset to stress test the hash table in isolation. We aim to prove the robustness of how it performs under different loading factors. As shown in Fig.~\ref{fig:hashtable}, it works seamlessly when the loading factor is under $85\%$. This is by far more than we need since the loading factor in operational mode never reaches more than $50\%$. The test performs 50 million insertions in the hash table, achieving a steady ten milliseconds for each insertion.

\begin{figure}[t]
    \centering
	\includegraphics[width=.7\columnwidth]{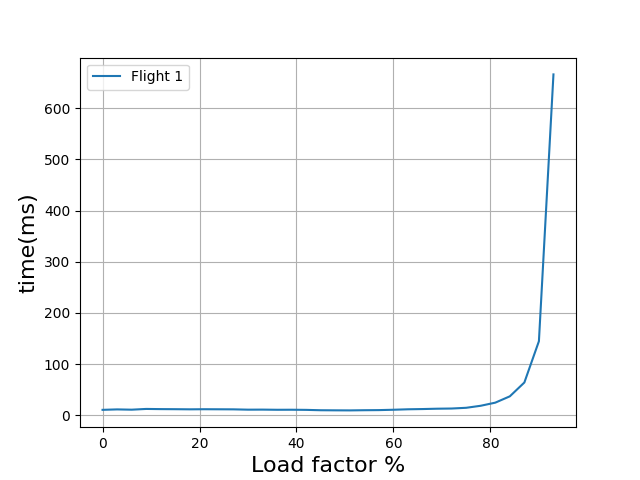}
	\caption{Execution time of Hash Table by loading factor, when inserting 50 million keys.}
	\label{fig:hashtable}
\end{figure}

\subsubsection{RMSE}

The RMSE error is computed from the trajectory of the robot and against the ground truth of the outdoor large-scale datasets. Table ~\ref{tab:rmse} shows similar errors for Voxgraph and coVoxSLAM running on PC. It also shows the accuracy of the system running on embedded devices like the Jetson Xavier AGX.

\setlength{\tabcolsep}{4pt}
\begin{table}[h!]
	\begin{center}
		\begin{tabular}{| c | c| c | c | c| } 
			\hline
			\multirow{2}{*}{\backslashbox{Flight}{Version}} & \multicolumn{2}{c|}{Standard Desktop PC}  & \multicolumn{2}{c|}{Nvidia Jetson Xavier} \\ 
			\cline{2-5}
			& Voxgraph & coVoxSLAM & Voxgraph & coVoxSLAM \\		
			
			\hline
			1 & 0.93 & \textbf{0.89} & --- & \textbf{1.20}\\
			\hline
			2 & \textbf{0.64} & 0.92 & --- & \textbf{0.97} \\
			\hline
			3 & \textbf{0.82} & 1.14 & --- & \textbf{1.21}\\
			\hline
			4 & \textbf{0.79} & 1.02 & --- & \textbf{1.03} \\
			\hline
		\end{tabular}
	\end{center}
	\caption{Root Mean Square Error (in meters) of the Absolute Trajectory Error (ATE) for Voxgraph and coVoxSLAM on different processing units.}
	\label{tab:rmse}
\end{table}

\section{CONCLUSIONS}
In this paper, we present a novel GPU-accelerated system coined coVoxSLAM, for building globally consistent volumetric maps in real time. Both the frontend and backend are designed to run fully on GPU to maximize efficiency and avoid conflicting unnecessary data transfers between CPU and GPU.
Our system implements the pose graph optimization algorithm to solve the backend's least square minimization problem on the GPU, and the results show that the system improves the running time execution while maintaining the same accuracy.
The system is optimized for operation on both discrete and embedded GPUs. We provide experiments demonstrating that coVoxSLAM is faster than other state-of-the-art approaches, like the recently released library nvBlox, achieving a speed-up of $1.5\times$ to $2\times$. Our raycasting merging algorithm outperforms current projective mapping alternatives while facilitating the adoption of better options for weighting techniques for the TSDF Integrator. We have released the source code of our system on GitHub, with examples and instructions to facilitate the use by the community and to ensure the repeatability of the shown experiments.

\section{ACKNOWLEDGMENT}
We want to thank the authors of Voxgraph, specifically Victor Reijgwart, for sharing the Voxgraph dataset.
\newpage
\bibliographystyle{IEEEtran}
\bibliography{IEEEabrv, biblio}

\end{document}